\documentclass[runningheads]{llncs}
\usepackage[T1]{fontenc}
\usepackage{newtxtext}       
\usepackage[varvw]{newtxmath}       

\usepackage{graphicx}

\usepackage{libertine}
\usepackage{amsmath}
\usepackage{mathtools}
\usepackage{bbold}
\usepackage{tikz-cd}
\usepackage{float}
\usepackage{wrapfig}
\usepackage[utf8]{inputenc}
\usepackage[
	colorlinks=true,
	pdfstartview=FitV,
	linkcolor=blue,
	citecolor=blue,
	urlcolor=blue
]{hyperref}
\usepackage{enumerate}
\usepackage{subcaption}
\usepackage{algorithm}
\usepackage{algpseudocode}

\numberwithin{equation}{section}
\numberwithin{figure}{section}
\DeclareFontFamily{U}{mathx}{}
\DeclareFontShape{U}{mathx}{m}{n}{<-> mathx10}{}
\DeclareSymbolFont{mathx}{U}{mathx}{m}{n}
\DeclareMathAccent{\wh}{0}{mathx}{"70}
\DeclareMathAccent{\wc}{0}{mathx}{"71}

\newcommand{\bitem}{\begin{itemize}}
\newcommand{\eitem}{\end{itemize}}
\newcommand{\mc}[1]{\mathcal{#1}}
\newcommand{\mb}[1]{\mathbb{#1}}

\newcommand{\N}{\mathbb{N}}
\newcommand{\R}{\mathbb{R}}

\newcommand{\V}{\mathbb{V}}

\newcommand{\bpm}{\begin{pmatrix}}
\newcommand{\epm}{\end{pmatrix}}
\newcommand{\bvm}{\begin{vmatrix}}
\newcommand{\evm}{\end{vmatrix}}
\newcommand{\bsm}{\left(\begin{smallmatrix}}
\newcommand{\esm}{\end{smallmatrix}\right)}
\newcommand{\T}{\top}

\newcommand{\wt}[1]{\widetilde{#1}}
\newcommand{\la}{\langle}
\newcommand{\ra}{\rangle}

\newcommand{\eins}{\mathbb{1}}

\newcommand{\LG}[1]{\mathrm{#1}}

\newcommand{\expm}{\exp_{\mathrm{m}}}

\DeclareMathSymbol{\mydiv}{\mathbin}{symbols}{"04}

\DeclareMathOperator{\tr}{tr}
\DeclareMathOperator{\Diag}{Diag}

\DeclareMathOperator{\rint}{rint}

\newcommand{\sst}[1]{{\scriptscriptstyle #1}}

\begin{document}

\title{Controlled Binary Assignment Flows}

\author{\small  Paul Fr\"{o}hling \inst{1}, Christoph Schn\"{o}rr \inst{1,2}}
\institute{Institute for Mathematics, Image and Pattern Analysis Group, Heidelberg University, Germany \and  Research Station Geometry and Dynamics, Heidelberg University}
\authorrunning{}

\maketitle              
\begin{abstract}
This paper introduces a geometric approach to the representation and learning of time-variant parametrizations of deep assignment flows, for binary data labelings on graphs. Each edge carries an optimally controlled parameter flow in a 2-dimensional totally geodesic submanifold, based on objectives derived from a non-local linear scale space representation of the input data. The aggregation of triplets of such flows in 5-dimensional totally geodesic submanifolds is learned from data. The fusion of these flows determines the trajectory in the high-dimensional symmetric parameter space and by geometric integration the layers of binary deep assignment flows. Numerical results demonstrate the expressivity and the limitation of our highly parallel geometric local synthesis approach to high-dimensional deep assignment flow parametrizations.

\end{abstract}

\section{Introduction}\label{sec:Introduction}

\textit{Geometric deep learning} \cite{Bronstein:2021tp} constitutes an active and promising research direction. The fast growing literature includes  works with a focus on manifold-valued data \cite{Bronstein:2017aa,Hanik:2026aa}, on the equivariance properties of mappings realized by deep networks \cite{Cohen:2016aa,Smets:2023aa,Gerken:2023aa} or on the application-specific sensitivity to spatial structure in terms of geometric control through sub-Riemannian geometry \cite{Sommer:2017aa,Bellard:2023aa}. Apparently, less work has been devoted to studying the \textit{geometric role of the network parameters themselves} for particular classes of deep networks.

This paper takes a step along this direction for the \textit{specific} problem class of \textit{binary} data labeling on \textit{arbitrary} graphs. The basis are \textit{assignment flows} \cite{Astrom:2017ac,Savarino:2021wt} of the form  
\begin{subequations}\label{eq:AF}
\begin{align}\label{eq:AF-dot-S}
\dot S(t) &= R_{S(t)}[\Omega(t) S(t)],\quad
S(0) = S_{0}\in\mc{S}_{c}^{n},
\quad\text{where}
\\
S &= (S_{1},\dotsc, S_{n})^{\T}\in\mc{S}_{c}^{n},
\\
\mc{S}_{c} 
&:= \rint(\Delta_{c}) = \{s\in\Delta_{c}\colon s>0\}
\quad(\text{rel.~interior:} 
&&\text{prob.~simplex})
\\
\mc{S}_{c}^{n} &:= \mc{S}_{c} \times\dotsb\times \mc{S}_{c} 
&&(\text{$n$ factors})
\\
(R_{S}[\Omega S])_{i} 
&:= R_{S_{i}} (\Omega S)_{i}
\\
&:= \Diag(S_{i})(\Omega S)_{i} - \la S_{i}, (\Omega S)_{i}\ra S_{i},\quad i\in[n],
&& (\text{row index})
\intertext{with \textit{time-variant symmetric positive-definite} parameter matrix}\label{eq:def-Om-general}
\Omega(t) &\in\mc{P}^{n}
:= \{A\in\R^{n\times n}\colon A=A^{\T},\; A\succ 0\},\quad t\geq 0.
\end{align}
\end{subequations}
The latter constraint satisfies a sufficient condition that the flow \eqref{eq:AF-dot-S} converges to integral labelings \cite{Zern:2020aa}.

\begin{figure}[]
    \centering
    \begin{minipage}[c]{0.568\textwidth}
        \centering
        \includegraphics[width=\linewidth]{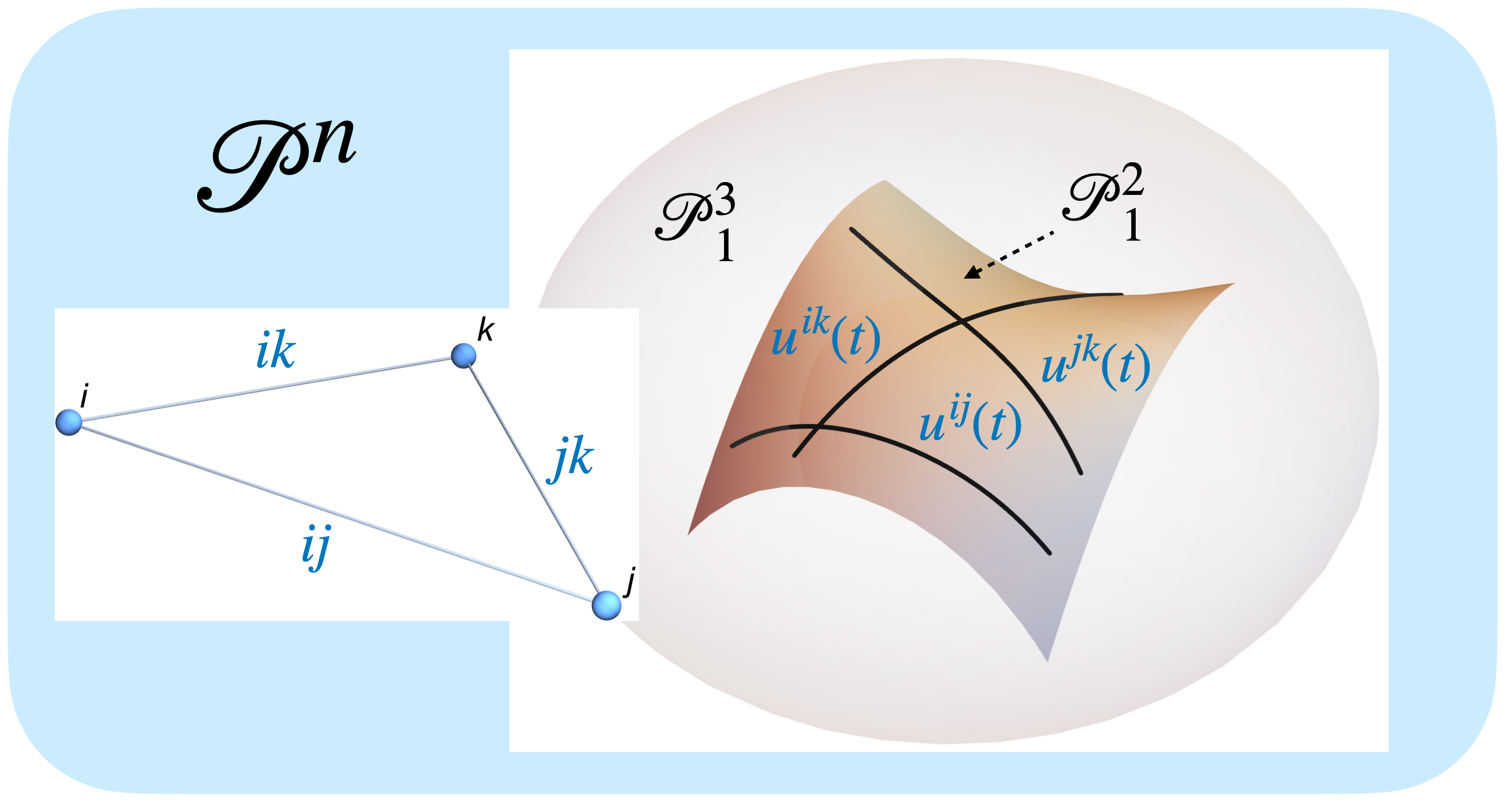}
    \end{minipage}
    \hspace{0.01 \textwidth}
    \begin{minipage}[c]{0.41\textwidth}
        \vspace{0.05 \textwidth}
        \caption{
        Illustration of our geometric local-to-global synthesis
        approach of high-dimensional parameter trajectories
        \eqref{eq:def-Om-general} for deep binary assignment flows,
        via the optimal control of trajectories and their
        aggregation in totally-geodesic submanifolds
        $\mc{P}_{1}^{2}, \mc{P}_{1}^{3}$.
        }
        \label{fig:intro}
    \end{minipage}
\end{figure}

Figure \ref{fig:intro} illustrates the first two stages of our \textit{3-stage approach} for \textit{representing and learning from data} the parameter trajectory \eqref{eq:def-Om-general} in the high-dimensional \textit{symmetric space $\mc{P}^{n}$}: \textbf{(1)} each edge $ij$ of the underlying graph indexes a control parameter trajectory $u^{ij}(t)$ in the 2-dimensional \textit{totally geodesic submanifold} $\mc{P}_{1}^{2}$ of positive-definite symmetric $2\times 2$ matrices with determinant $1$, which are \textit{globally optimized} in parallel on all edges through discrete optimal control. \textbf{(2)} Triplets of such flows indexed by \textit{triangular subgraphs} are \textit{aggregated via machine learning} to form time-variant $3\times 3$ parameter matrices in the 5-dimensional totally geodesic submanifold $\mc{P}_{1}^{3}$. \textbf{(3)} The third ingredient (not shown in Fig.~\ref{fig:intro}) is a small finite subset of the \textit{nonlocal linear scale space representation} of the input data, used to define the objectives of discrete optimal control processes at stage (1). 

Altogether, vertices $\mc{V}=\{\dotsc,i,j,k,\dotsc\}$ and edges $\mc{E}=\{\dotsc,ij, ik, jk,\dotsc\}$ of the underlying graph $\mc{G}=(\mc{V},\mc{E})$ determine a map $u(t)\mapsto \Omega(t)\in\mc{P}^{n}$ and by geometric integration \cite{Zeilmann:2020aa} of \eqref{eq:AF-dot-S} the resulting deep assignment flow network.

Section \ref{sec:Binary-AF} specializes the general set-up \eqref{eq:AF} to binary labelings. The ingredients (1), (2), (3) of our approach sketched above are introduced in Section \ref{sec:Approach}. Numerical results illustrate in Section \ref{sec:Experiments} both the expressivity and the limitation of our geometric approach to the parametrization of high-dimensional binary labeling via assignment flows. Our work paves the way for future research indicated in Section \ref{sec:Conclusion}.

\section{Binary Assignment Flows}\label{sec:Binary-AF}

We consider the specific case of two labels $c=2$, i.e.~binary data labeling, and reformulate the general assignment flow equation \eqref{eq:AF} accordingly. Since $\Delta_{2}\cong [0,1]$, any probability vector $p = (p_{1},1-p_{1})\in\Delta_{2}$ is determined just by its first component, so is componentwise any assignment vector $S\in\mc{S}_{2}^{n}$. As a result, we have the following.
\begin{proposition}[\textbf{binary assignment flow equation}]\label{prop:binary-AF}

Let $c=2$ and define the map 
\begin{equation}
s\colon\mc{S}_{2}^{n}\to (0,1)^{n},\qquad
s(S) = (s_{1},\dotsc,s_{n})^{\T}
:= (S_{11},\dotsc, S_{n1})^{\T}.
\end{equation}
Then the assignment flow equation \eqref{eq:AF} is equivalent to the system
\begin{subequations}\label{eq:def-binary-AF}
\begin{align}\label{eq:def-binary-AF-S-s}
S(t) &= \left(\bpm \sigma\big(2 v_{1}(t)\big) \\ 1 - \sigma\big(2 v_{1}(t)\big) \epm, \dotsc, 
\bpm \sigma\big(2 v_{n}(t)\big) \\ 1 - \sigma\big(2 v_{n}(t)\big) \epm\right)^{\T}
\\
&=: s^{-1}\big(\sigma[2 v(t)]\big),
\intertext{where $\sigma(x) = (1 + e^{-x})^{-1}$ is the logistic sigmoid function and
$v(t) = \big(v_{1}(t),\dotsc, v_{n}(t)\big)^{\T} \in\R^{n}$ solves} \label{eq:def-binary-AF-v-dot}
\dot v &= \frac{1}{2} \Omega \tanh[v],\qquad
v(0) = v_{0} := \frac{1}{2}\log\Big[\frac{(S_{0})_{,1}}{\eins_{n}-(S_{0})_{,1}}\Big] \in \R^{n}.
\end{align}
\end{subequations}
\end{proposition}
As a consequence of Proposition \ref{prop:binary-AF}, it suffices to solve the nonlinear system of ODEs \eqref{eq:def-binary-AF-v-dot} in order to determine the binary label assignment by substitution into \eqref{eq:def-binary-AF-S-s} followed by trivial rounding.

The binary assignment flow network corresponding to \eqref{eq:def-binary-AF} is derived by basic geometric integration. Following \cite{Zeilmann:2020aa} and 
using a discrete layer index also denoted by $t\in [T] = [1,2,\dotsc,T],\; T\in\N$, the application of the explicit geometric Euler scheme with step sizes $h_{t}$ yields the
\begin{subequations}\label{eq:def-binary-AF-network}
\begin{align}
S_{t+1} &= s^{-1}(\sigma[2 v_{t+1}]),
&&(\textbf{binary assignment flow network})
\label{eq:def-binary-AF-network-S} 
\\ 
\label{eq:def-binary-AF-network-v}
v_{t+1} &= v_{t} + h_{t} \Omega_{t} \tanh[v_{t}],
\quad
t \in [T],
&& \Omega_{t}\in\mc{P}^{n},
\end{align}
\end{subequations}
where the step sizes $h_{t}$ absorb the factor $\frac{1}{2}$ from \eqref{eq:def-binary-AF-v-dot}. The network \textit{input} is $v_{1}$, the network \textit{output} is $S_{T+1}$. Inspecting \eqref{eq:def-binary-AF-network} reveals the \textit{basic network structure}:
\begin{itemize}
\item each network layer $t\in [T]$ comprises a $\tanh$ activation function and a \textit{manifold-valued} parameter matrix $\Omega_{t}$ corresponding to \eqref{eq:def-Om-general};
\item each layer $t \in [T+1]$ outputs the state $S_{t}$, including the initial layer with output equal to the input $S_{1}$ encoded by input vector $v_{1}$.
\end{itemize}
We point out that this structure is a \textit{consequence} of the information geometry underlying the assignment flow equation \eqref{eq:AF}, rather than set up beforehand in an ad-hoc manner. 
In Section \ref{sec:Approach}, we study the problem to learn the parameters $\Omega_{t}\in\mc{P}^{n},\; t\in[T]$ 
of the network \eqref{eq:def-binary-AF-network}.

\section{Approach}\label{sec:Approach}
This section details our approach illustrated by Figure \ref{fig:intro}. Each of the three stages is described below in a corresponding subsection.

\subsection{Optimal Parameter Trajectories on Edges}\label{sec:Approach-2x2}

We consider the \textit{restriction} of the discrete-time flow \eqref{eq:def-binary-AF-network} to the edges $ij\in\mc{E}$ as minimal \textit{basic components} of assignment flow networks and call them \textit{$2 \times 2$ systems}.
Each $2 \times 2 $ systems couples two discrete binary probability distributions. 
For each $t$ and edge $ij$, we impose the constraint
\begin{equation}\label{eq:Om-ij-constraint}
\Omega^{(ij)}_{t}\in\mc{P}_{1}^{2} 
:= \{A\in\mc{P}^{2}\colon \det(A)=1\},\qquad \forall t\in [T],\quad
\forall ij\in\mc{E}(\mc{G}),
\end{equation}
in terms of the totally geodesic submanifold $\mc{P}_{1}^{2}$ of the positive definite manifold \eqref{eq:def-Om-general} for $n=2$. 
We take this constraint into account by the tangent space parametrization, utilizing the matrix exponential $\exp_m$ as global diffeomorphism
\begin{subequations}\label{eq:def-u-ij-T}
\begin{align}
\mb{S}^{2}_{0} 
:= T_{\Omega_{t}}\mc{P}_{1}^{2} \ni u^{(ij)}_{t} 
&= \bpm u_{i} \\ u_{j} \epm_{t}
\;\mapsto\;
\bpm u_{i} & u_{j} \\  u_{j} & -u_{i} \epm_t
\xrightarrow{\expm}
\Omega_{t}^{(ij)} = \Omega(u^{(ij)}_{t})
\in\mc{P}_{1}^{2}
\\
\mb{S}^{2}_{0}
&:= \{U\in\R^{2\times 2} \colon U^{\T}=U,\; \tr U=0\}.
\end{align}
\end{subequations}
Using the shorthands
\begin{subequations}\label{eq:def-u-v-T}
\begin{align}
u_{\sst{[T]}} &:= (u_{t})_{t\in[T]} = (u_{1},\dotsc, u_{T}),\label{eq:def-u-v-T-u} \\ \label{eq:def-u-v-T-v}
v_{\sst{[T+1]}} &:= (v_{t})_{t\in[T+1]} = (v_{1},\dotsc, v_{T+1}),
\end{align}
\end{subequations}
we call \textit{optimal} a sequence $u_{\sst{[T]}}$ if it minimizes the objective function
\begin{equation}\label{eq:J-u-general}
J(u_{\sst{[T]}}, v_{\sst{[T+1]}}) 
= \sum_{t\in[T]}\ell_{t+1}(v_{t}, u_{t})
\end{equation}
given in terms of loss functions $\ell_{2},\dotsc,\ell_{T+1}$ to be specified. 

\begin{figure}
    \centering
    \begin{minipage}[c]{0.65\textwidth}
        \centering
        \resizebox{\linewidth}{!}{%
        \begin{tikzpicture}
            \node[anchor=west] at (0,6)
                {\textbf{local binary assignment flow}};

            \draw[->] (1,3.3) -- (4,3.3);
            \draw[->] (1.3,3) -- (1.3,5);

            \node[circle,fill=gray,inner sep=2pt]
                (vt) at (2.1,4.6) {};
            \node[right] at (2.2,4.7) {$v_t$};

            \node[circle,fill=black,inner sep=2pt]
                (vnext) at (2.5,4.3) {};

            \draw[->,thick] (vt) -- (vnext);

            \node[right] at (2.7,4.3)
                {$v_{t+1}=v_t+h_t\Omega_t\tanh(v_t)$};

            \draw[->,thick,dashed]
                (vnext) -- (2.5,1.92);

            \node[anchor=west] at (7.9,6)
                {\textbf{parametrization}};

            \node[
                draw,
                rounded corners,
                thick,
                minimum width=3.2cm,
                minimum height=1.2cm,
                align=center
            ] (U) at (10.1,4.2)
                {$U_t\in\mb{S}_{0}^{2},\;
                \mathrm{tr}(U_t)=0$};

            \node[
                draw,
                rounded corners,
                thick,
                minimum width=3.2cm,
                minimum height=1.0cm,
                align=center
            ] (Omega) at (10.1,1.4)
                {$\Omega_t\in\mathcal P_1^2$};

            \draw[->,thick,dashed]
                (U.south) -- (Omega.north)
                node[midway,right] {$\exp_m(U_t)$};

            \draw[->,thick,dashed]
                (8.5,1.8) -- (5.4,4.0);

            \node[right] at (2.5,2.6)
                {$\sigma(2v_{t+1})$};

            \node[
                draw,
                rounded corners,
                thick,
                minimum width=3.2cm,
                minimum height=1.0cm,
                align=center
            ] at (2.5,1.4)
                {$s_{t+1}\in\mc{S}^2_2$};

        \end{tikzpicture}%
        }
    \end{minipage}%
    \hfill
    \begin{minipage}[c]{0.3\textwidth}
        \vspace{0.09 \textwidth}
        \centering
        \includegraphics[width=\linewidth]{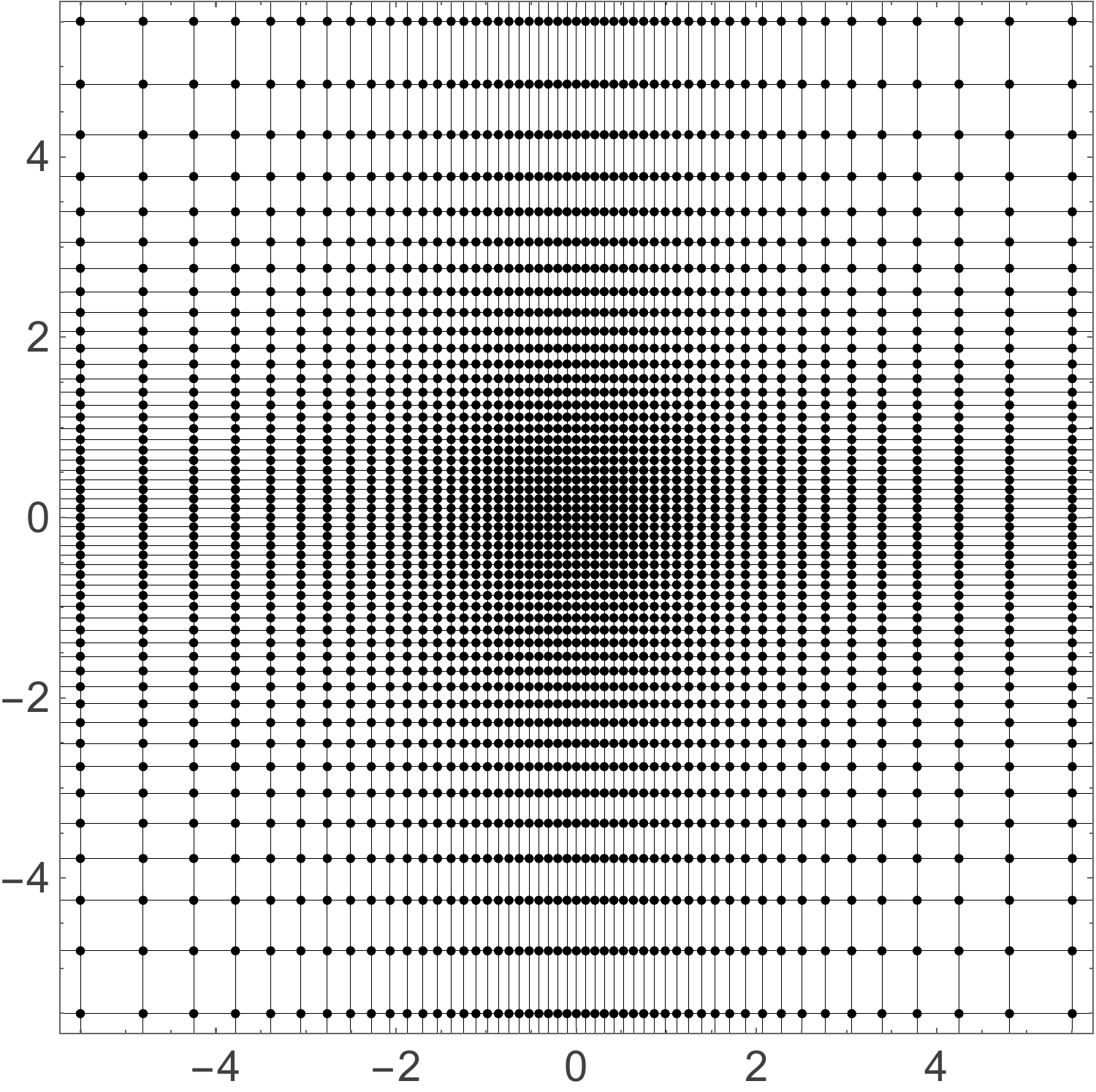}
    \end{minipage}

    \caption{
        \textbf{Left}: Illustration of the layer mapping and its
        parametrization of the local binary assignment network
        \eqref{eq:def-binary-AF-network-v},
        for an arbitrary edge $ij\in\mc{E}(\mc{G})$
        of the underlying graph (dropped above from the notation)
        and the corresponding pairs of input feature vectors
        and output assignment vectors.
        The control parameters \eqref{eq:def-u-ij-T}
        which determine the parameter matrix, determine in turn
        the flow.
        We learn local binary assignment flows by determining
        optimal trajectories of these parameters by discrete
        optimal control.
        \textbf{Right}: The feasible set for vectors of the sequences \eqref{eq:def-u-v-T-v}. Regions near the origin  corresponding to a large uncertainty of \textit{local} label assignments and hence are discretized more finely, whereas regions farther from the origin are discretized more coarsely.
    }
    \label{fig:assignment-flow}
\end{figure}
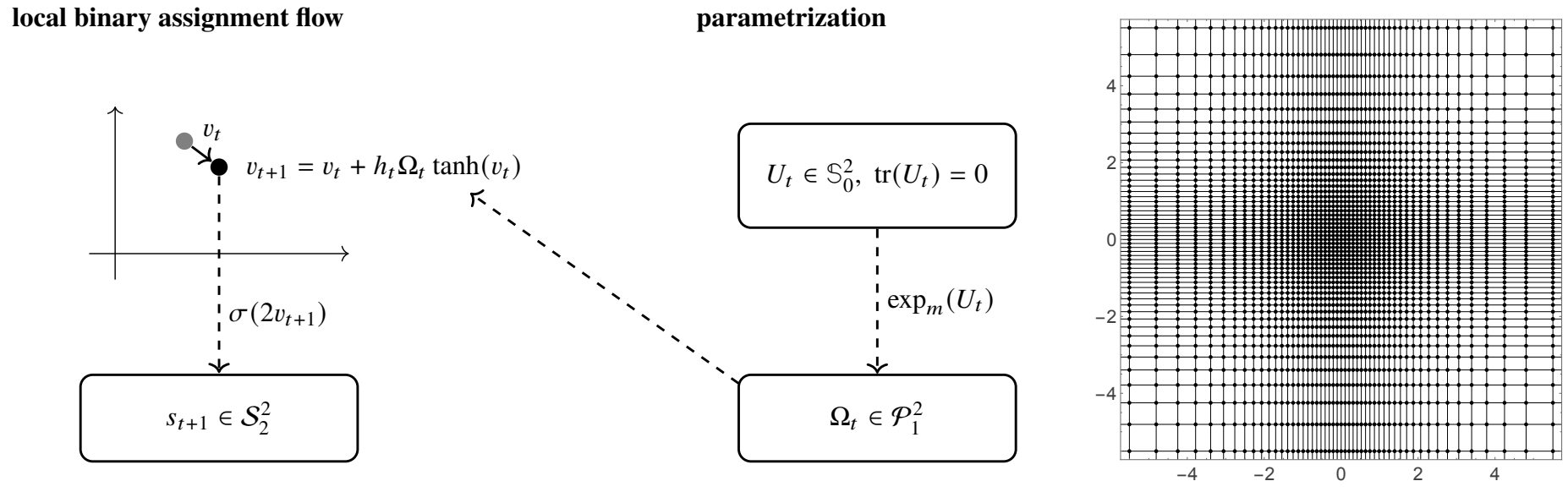

Determining optimal parameter sequences \eqref{eq:def-u-v-T-u} is accomplished by dynamic programming \cite{HernandezLerma:2023}.  
Consider the minimization problem \eqref{eq:J-u-general} for any given initial vector $v_{1}\in\R^{2}$ and a target vector $v_{T+1}$ corresponding to a prescribed target state. 
Let the \textit{cost-to-go functions} $\mathcal{J}_{[t,T]},\,t\in[T]$ be defined \textit{recursively backward} for $t = T, T-1, \dots, 1$ by 
\begin{subequations}\label{eq:def-Bellman}
\begin{align}
    \mathcal{J}_{[T,T]}(v_{T}) 
    &:= \min_{u_T \in \mathbb{U}^{2}} \ell_{T+1}(v_T, u_T) %
\intertext{and the \textit{Bellman equation}}\label{eq:def-Bellmann}
    \mathcal{J}_{[t,T]}(v_{t}) 
    &:= \min_{u_{t} \in \mathbb{U}^{2}} \big\{ \ell_{t+1}(v_{t}, u_{t}) + \mathcal{J}_{[t+1,T]}\big(\Phi(v_{t}, u_{t})\big) \big\},\quad
    t = T-1,\dotsc, 2, 1.
\end{align}
\end{subequations}
on the \textit{finite} feasible set $\V^2 \subset \R^2$ depicted by Fig.~\ref{fig:assignment-flow} for each vector $v_{t},\; t\in[T]$, 
and on a \textit{finite} regular grid $\mb{U}^2 \subset \R^2$ of feasible control parameter vectors of the sequence \eqref{eq:def-u-v-T-u}. The function $\Phi$ in \eqref{eq:def-Bellmann}, defined by 
\begin{equation}
\Phi(v_{t}, u_{t}) = \operatorname{clip}_{\mathbb{V}}\big(v_{t} + \Omega(u_{t})\tanh(v_{t})\big),\qquad
\operatorname{clip}_{\mathbb{V}}
\colon\R^{2}\to\mb{V}^{2}
\end{equation}
maps to the feasible set by rounding. 
For our implementation, we have chosen the weighted loss functions
\begin{align}
    \ell_{t+1}(v_t, u_t) := \lambda_t \Vert \Phi(v_t, u_t) - v_{T+1} \Vert ^2,\qquad t\in[T],
\end{align}
where the weights $\lambda_t>0$ \textit{increase} with $t$ in order to assign more importance to states closer to the terminal target.

Dynamic Programming solves the optimization problem \eqref{eq:J-u-general} by working recursively backward,  starting at the target.
This leads to the \textit{policy mapping} 
\begin{align}
    \label{eq:def-policy-pi}
    \pi_{v^{*}_{T+1}} \colon \mathbb{V}^{2} \times [T] \to \mathbb{U}^{2}, \qquad (v_t, t) \mapsto u_{t}^*
\end{align}
returning an optimal $u_t^{*}$ for a current state $(v_t, t)$ and a given target $v_{T+1}$. 
If, for each $t\in[T]$, the optimality of $u_{t}^{\ast}$ is achieved in \eqref{eq:def-Bellman}, then the entire sequence $u_{[T]}^* = (u_1^*, \dots, u_{T}^*)$ given by \eqref{eq:def-policy-pi} is optimal.
Using the resulting policy $\pi_{v_{T+1}^*}$, optimal trajectories emanating from any feasible point can be determined for a given target $v_{T+1}^*$.

\subsection{Learning Flow Aggregation on Triangular Subgraphs}\label{sec:Approach-3x3}
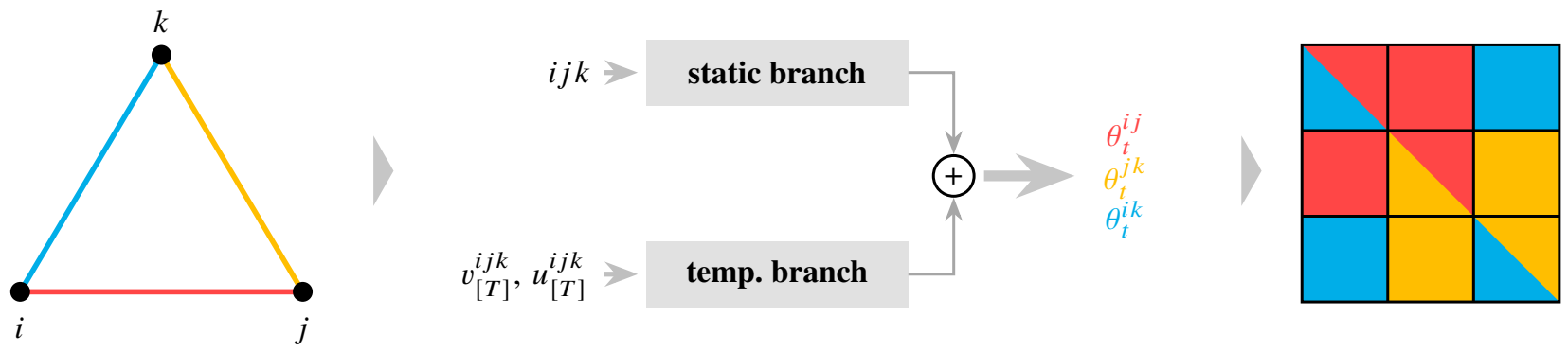
\begin{figure}
    \centering
     \resizebox{\textwidth}{!}{
    \usetikzlibrary{
    arrows.meta,
    positioning,
    calc,
    shapes.geometric
}

\definecolor{myred}{RGB}{255,70,70}
\definecolor{myblue}{RGB}{0,174,232}
\definecolor{myyellow}{RGB}{255,190,0}
\definecolor{mygray}{RGB}{215,215,215}

\begin{tikzpicture}[
    >=Stealth,
    font=\sffamily,
    thick
]

\begin{scope}[xshift=-2.5cm]

\coordinate (i) at (0.2,2.2);
\coordinate (j) at (3.0,2.2);
\coordinate (k) at (1.6,4.55);

\draw[myred,line width=1.5pt] (i)--(j);
\draw[myblue,line width=1.5pt] (i)--(k);
\draw[myyellow,line width=1.5pt] (k)--(j);

\foreach \p in {i,j,k}
    \fill (\p) circle (2.8pt);

\node[below=4pt] at (i) {$i$};
\node[below=4pt] at (j) {$j$};
\node[above=3pt] at (k) {$k$};

\end{scope}

\begin{scope}[xshift=-2.6cm]
    \fill[gray!45]
        (3.8,3.05) --
        (4.0,3.4) --
        (3.8,3.75) -- cycle;
\end{scope}

\begin{scope}[
    rotate around={90:(6.1,3.0)},
    xshift=0.35cm,
    yshift=0.5cm,
    transform shape
]

\node[rotate=-90] at (5.1,5.9)
    {$v^{ijk}_{[T]},\,u^{ijk}_{[T]}$};

\node[rotate=-90] at (7.1,5.45) {$ijk$};

\fill[mygray!75] (4.8,2.1) rectangle (5.45,4.7);
\fill[mygray!75] (6.8,2.1) rectangle (7.45,4.7);

\node[rotate=-90,font=\bfseries]
    at (5.125,3.4) {temp. branch};

\node[rotate=-90,font=\bfseries]
    at (7.125,3.4) {static branch};

\draw[-Stealth,gray!50,line width=1.4pt]
    (5.125,5.12)--(5.125,4.78);

\draw[-Stealth,gray!50,line width=1.4pt]
    (7.125,5.12)--(7.125,4.78);

\node[draw,circle,fill=white,
      inner sep=1.5pt] (plus)
    at (6.1,1.65) {$+$};

\draw[gray!70,-Stealth]
    (5.125,2.1)--(5.125,1.65)--(plus.west);

\draw[gray!70,-Stealth]
    (7.125,2.1)--(7.125,1.65)--(plus.east);

\draw[-Stealth,gray!45,line width=3pt]
    (6.1,1.35)--(6.1,0.45);

\node[rotate=-90, align=center] at (6.1,-0.05) {
    $\textcolor{myred}{\theta_t^{ij}}$ \\
    $\textcolor{myyellow}{\theta_t^{jk}}$ \\
    $\textcolor{myblue}{\theta_t^{ik}}$
};

\end{scope}

\begin{scope}[xshift=6cm]
    \fill[gray!45]
        (3.8,3.05) --
        (4.0,3.4) --
        (3.8,3.75) -- cycle;
\end{scope}-----

\begin{scope}[shift={(10.4,2.1)}]
    \def\s{0.85}

    \fill[myblue]   (0,2*\s) rectangle (\s,3*\s);
    \fill[myred]    (\s,2*\s) rectangle (2*\s,3*\s);
    \fill[myblue]   (2*\s,2*\s) rectangle (3*\s,3*\s);

    \fill[myred]    (0,\s) rectangle (\s,2*\s);
    \fill[myyellow] (\s,\s) rectangle (2*\s,2*\s);
    \fill[myyellow] (2*\s,\s) rectangle (3*\s,2*\s);

    \fill[myblue]   (0,0) rectangle (\s,\s);
    \fill[myyellow] (\s,0) rectangle (2*\s,\s);
    \fill[myblue]   (2*\s,0) rectangle (3*\s,\s);

    \fill[myred]
        (0,3*\s)--(\s,3*\s)--(\s,2*\s)--cycle;

    \fill[myred]
        (\s,2*\s)--(2*\s,2*\s)--(2*\s,\s)--cycle;
    
    \fill[myyellow]
    (2*\s,\s)--(3*\s,\s)--(3*\s,0)--cycle;

    \draw[line width=0.8pt]
        (0,0) rectangle (3*\s,3*\s);

    \foreach \x in {1,2}
        \draw (\x*\s,0)--(\x*\s,3*\s);

    \foreach \y in {1,2}
        \draw (0,\y*\s)--(3*\s,\y*\s);

\end{scope}

\end{tikzpicture}
     }
    \caption{Left: A triangle (minimally complete) subgraph. Each edge indexes an optimally controlled binary assignment flow network \eqref{eq:def-binary-AF-network-v}. Fusing these minimal networks into a larger network creates conflicts because each controlled variable participates in two local systems, as indicated by the above right-most panel and the application of the injection map \eqref{eq:def-iota-u-ij} to the three sequences \eqref{eq:t-intro-uvw-T}. This problem is resolved by a machine learning component (center panel) with learnable parameters $\theta_{t}^{e},\, e\in\{ij, ik, jk\}$.}
    \label{fig:ijk}
\end{figure}

%
In this section, as sketched by Figure \ref{fig:ijk}, we describe the aggregation of $3\times3$ flows on \textit{triangular} subgraphs based on three local $2\times2$ flows on the corresponding edges, each determined via dynamic programming as described in Section \ref{sec:Approach-2x2}.

\textbf{Geometric Ansatz.} We consider any triangle $T_{ijk}$ and denote the vertices $\{i,j,k\}$ by $\{1,2,3\}$, for simplicity. For the edges $ij=12$, $ik=13$ and $jk=23$, we denote the optimal parameter sequences by
\begin{equation}\label{eq:t-intro-uvw-T}
u^{(12)}_{[T]},\quad
u^{(13)}_{[T]},\quad
u^{(23)}_{[T]}.
\end{equation}
The control parameters of the $2 \times 2$ edge flows are embedded into $3 \times 3$ parameter matrices via the map $\iota\colon\R^{2}\to\mb{S}_{0}^{3}$, 
\begin{equation}
\label{eq:def-iota-u-ij}
\iota(u^{(12)}_t)=\bpm
u_{1} & u_{2} & 0 \\
u_{2} & -u_{1} & 0\\
0      & 0      &0
\epm_t, \quad 
\iota(u^{(13)})=\bpm
u_{1} & 0 & u_{3}  \\
0      & 0 &0\\
u_{3}  & 0 & -u_{1} 
\epm_t, \quad 
\iota(u^{(23)})=\bpm
0 & 0      & 0 \\    
0 & u_{2}& u_{3} \\
0 & u_{3} & -u_{2}
\epm_t.
\end{equation}

The image of the parameter space $\R^{2}$ under the composition of this embedding and the exponential map,
\begin{equation}
\exp_{m}\circ\iota(\R^{2}) \subset \mc{P}_{1}^{3},
\end{equation}
forms a \textit{totally geodesic submanifold} in the manifold $\mc{P}_{1}^{3}$, since the linear subspace $\iota(\R^{2})$ forms a Lie triple system 
\cite{Helgason:1978aa,Tumpach:2024aa}: evaluating the double-commutator on matrices parametrized by arbitrary
 parameter vectors $u^{(12)}_{t}, v^{(12)}_{t}, w^{(12)}_{t}\in\R^{2}$ yields again a matrix in the submanifold,
\begin{equation}
\big[\iota(u^{(12)}_{t}),[\iota(v^{(12)}_{t}),\iota(w^{(12)}_{t})]\big] \in \iota(\R^{2}),\qquad
\forall u^{(12)}_{t}, v^{(12)}_{t}, w^{(12)}_{t}\in\R^{2},
\end{equation}
i.e.~there is a vector $z^{(12)}_{t}$ depending on $u^{(12)}_{t}, v^{(12)}_{t}, w^{(12)}_{t}$ such that the double-commutator evaluates to $\iota(z^{(12)}_{t})$.  

\begin{figure}
\centerline{
\includegraphics[width=0.3\textwidth]{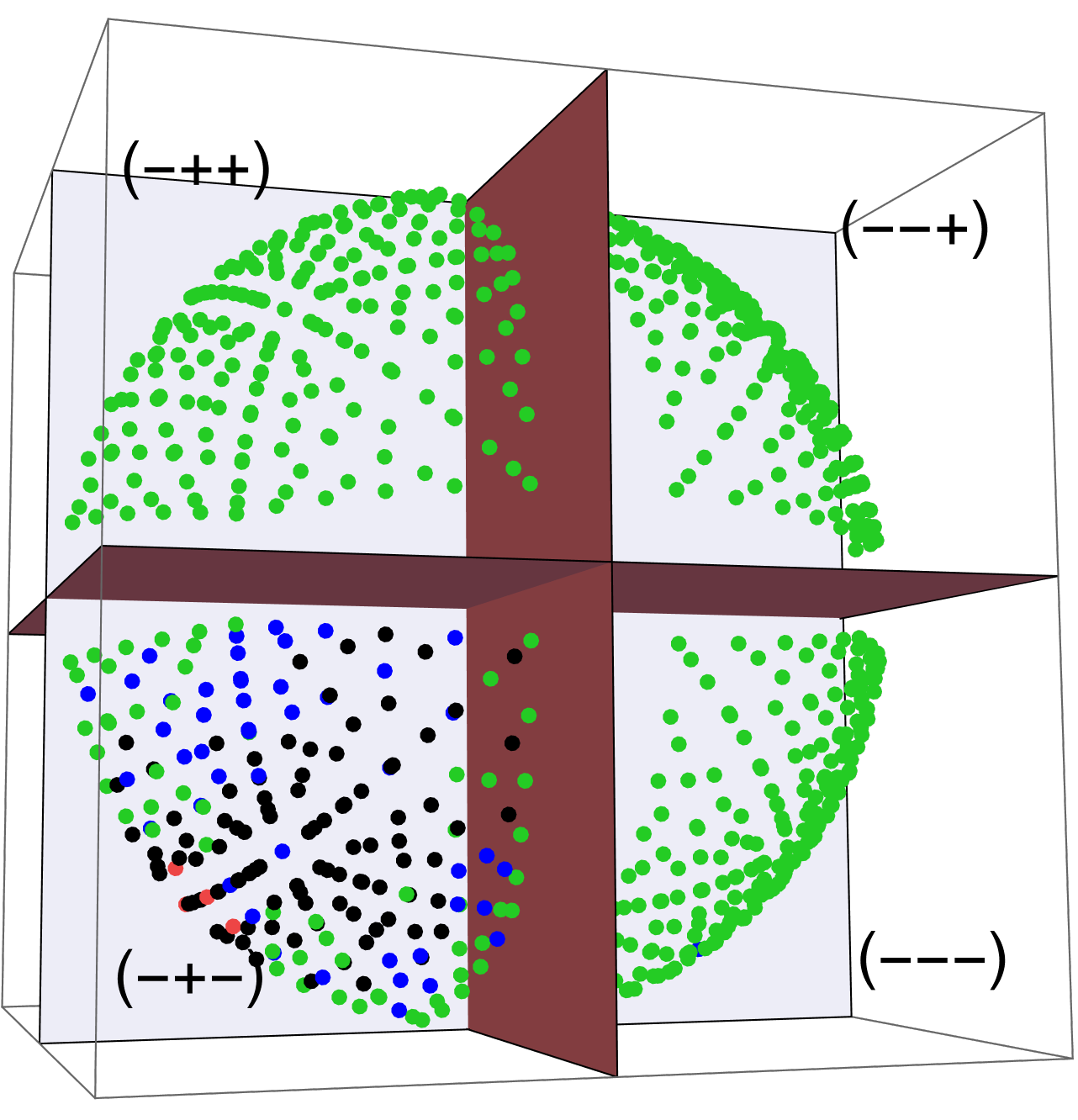}
\hspace{0.05\textwidth}
\includegraphics[width=0.3\textwidth]{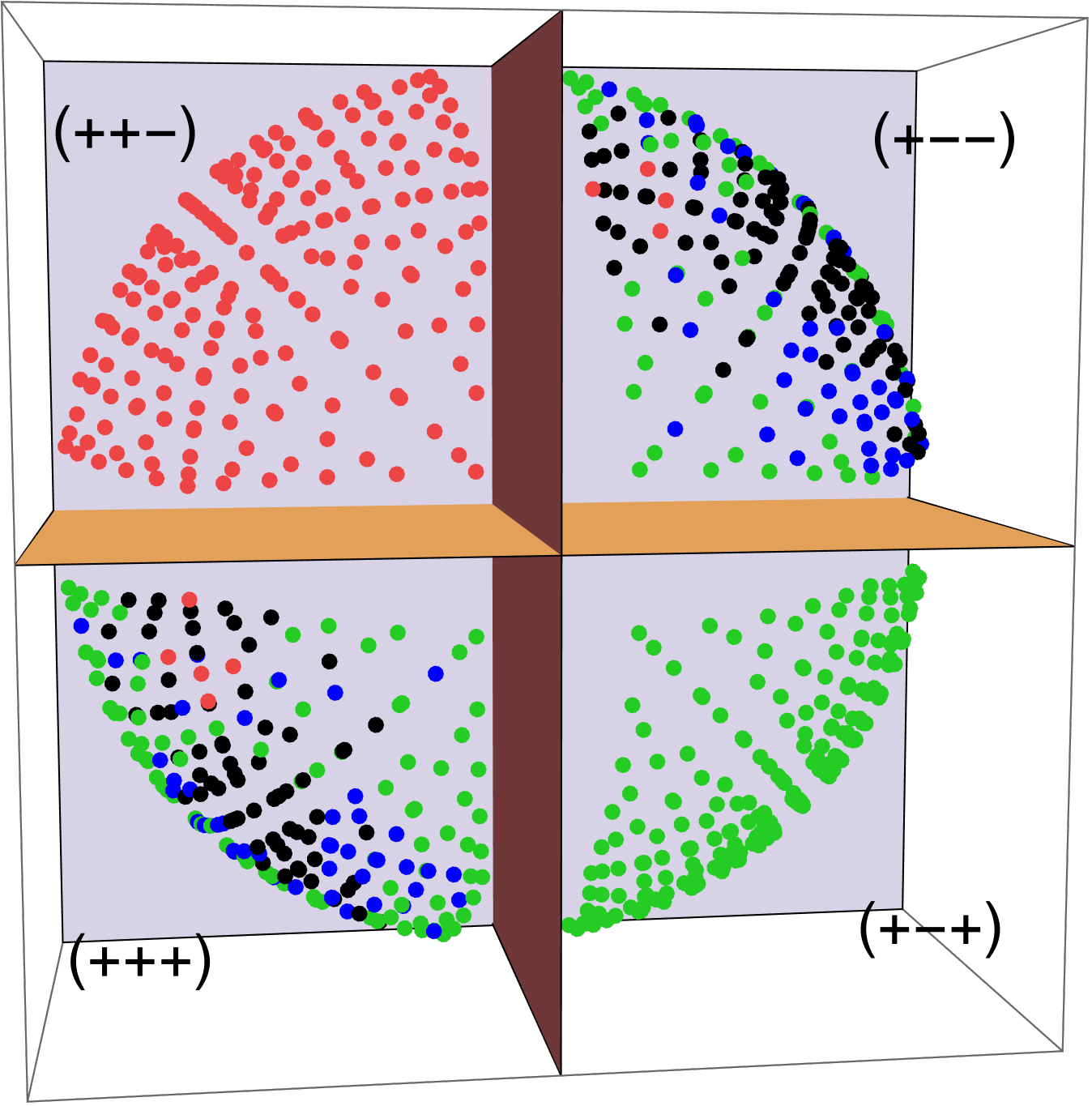}
}
\caption{
Performance of locally labeled triangular subgraphs, based on local geometric averaging \eqref{eq:plain-aggregation} of three $2\times 2$ flows indexed by the edges. Each point indicates an initial point around $0$, mapped to a small sphere for visualization. Its color displays if the trajectory \eqref{eq:def-binary-AF-network-v} reaches the target octant $(--+)$ (green) or makes a single (blue), two (black) or three (red) labeling errors at the vertices. Errors occur when the starting and target octants differ by $\geq 2$ binary labels.
}
\label{fig:3x3-Traj-Averaged}
\end{figure}

As a consequence and in view of the optimality of the parameter sequences \eqref{eq:t-intro-uvw-T}, it appears natural to define
the sequence $\Omega_{[T]}\subset\mc{P}_{1}^{3}$ for the triangular subgraph by averaging the tangent matrices corresponding to the three edges and subsequent exponentiation \cite{Arsigny:2007},
\begin{equation}\label{eq:plain-aggregation}
\mc{P}_{1}^{3} \supset \Omega_{[T]}\ni\Omega_{t}
:= \expm\Big(\frac{1}{2}\big(\iota(u_{t}^{(12)}) + \iota(u_{t}^{(13)}) + \iota(u_{t}^{(23)})\big)\Big),\quad t\in[T].
\end{equation}
For a numerical evaluation of this Ansatz, the starting points $\{v_{1}\in\mb{V}\colon \|v\|_{\infty}<0.7\}$ were selected and every vector $v_{T+1}$ in the orthant $(-,-,+)$ was considered as admissible target. 
For each pair $(v_{1},v_{T+1})$, three optimal edge flows were constructed that achieve the mappings $(v_{1;1}, v_{1;2}) \rightarrow (-,-)$, $(v_{1;1}, v_{1;3}) \rightarrow (-,+)$, $(v_{1;2}, v_{1;3}) \rightarrow (-,+)$ corresponding to the restrictions of the desired $3\times 3$ flow.
Empirically, it turned out that for the resulting set of 1728 test flows, the aggregation approach \eqref{eq:plain-aggregation} by plain geometric averaging  only achieved $\sim 62$\% correct $3\times 3$ flows, caused by local conflicts illustrated by Figure \ref{fig:ijk}, right panel.

\textbf{Additional learning component.} We refine the Ansatz \eqref{eq:plain-aggregation} by replacing each vector $u_{t}^{(ij)}$ by the vectors 
\begin{equation}\label{eq:uij-R}
    \wt{u}^{(ij)}_t= R(\theta^{(ij)}_t) u^{(ij)}_t, \qquad 
    R(\theta^{(ij)}_t) = \bpm \cos(\theta^{(ij)}_t), -\sin(\theta^{(ij)}_t) \\ \sin(\theta^{(ij)}_t), \cos(\theta^{(ij)}_t) \epm\in\LG{SO}(2),
\end{equation}
each adjusted by a planar time-variant rotations with learnable parameter $\theta^{(ij)}_{t}\in[0,2\pi)$. These parameters are learned by a neural network which realizes the mapping
\begin{equation}\label{eq:network-F}
    F_{\theta}(\xi) := (\theta^{(ij)}_t, \theta^{(ik)}_t, \theta^{(jk)}_t), \qquad \xi=\{\xi_1, \xi_2,...,\xi_n\}.
\end{equation}
This mapping is used not only to \textit{align} local edge flows in order to avoid to labeling errors displayed by Figure \ref{fig:3x3-Traj-Averaged}, but also to \textit{reorient} flows, provided the \textit{non-local} context (cf.~Section \ref{sec:Approach-ScaleSpace}) suggests to change the target labeling at some point of time $t\in[T]$.

The architecture of the neural network consists of two branches (Fig.~\ref{fig:ijk}, center panel): one for \textit{static} features and one for \textit{time-dependent} features.
The static branch is a three-layer neural network. The temporal branch consists of an LSTM \cite{Yu:2019aa} for processing the $u$ and $v$- trajectories of the three edge flows of a triangular subgraph.

\vspace{0.1cm}
\noindent
As input $\xi$ of the network \eqref{eq:network-F}, we use
\begin{itemize}
    \item \textbf{static features}: initial pixel positions and values $v_{1}$ per incident triangular subgraph;
    \item \textbf{temporal features}: $v$-trajectories, $u$-trajectories for every edge, current time (layer index) $t$, current state in $\R^3$ (starting point $v_{1}$ or latest prediction $\wt{v}_{t}$ for $t>1$).
\end{itemize}
Minimizing the loss function ($m$: margin, $\mc{T}$: set of triangles)
\begin{equation}\label{eq:loss-function}
    \mathcal{L}(\xi) 
                             =  \frac{1}{|\mc{T}|}\sum_{\tau \in \mathcal{T}}\sum_{i\in\tau} \log\Big(1+ \exp\big(m - \operatorname{sgn}(v^*_{i;T+1}) \wt{v}_{i;T+1}(\xi)\big)\Big) \\ 
\end{equation}
with margin parameter $m>0$ and dependency $\xi\mapsto \wt{v}_{i;T+1}(\xi)$ given by \eqref{eq:def-binary-AF-network-v}, \eqref{eq:uij-R} and \eqref{eq:network-F}, 
enforces that each aggregated flow for a triangle reaches the prescribed octant. After training, repeating the local labeling of triangular subgraphs (Fig.~\ref{fig:3x3-Traj-Averaged}) achieves $100$\% performance.

\subsection{Context Representation via a Linear Scale Space}
%

\begin{figure}
    \centering
    \includegraphics[width=0.65\linewidth]{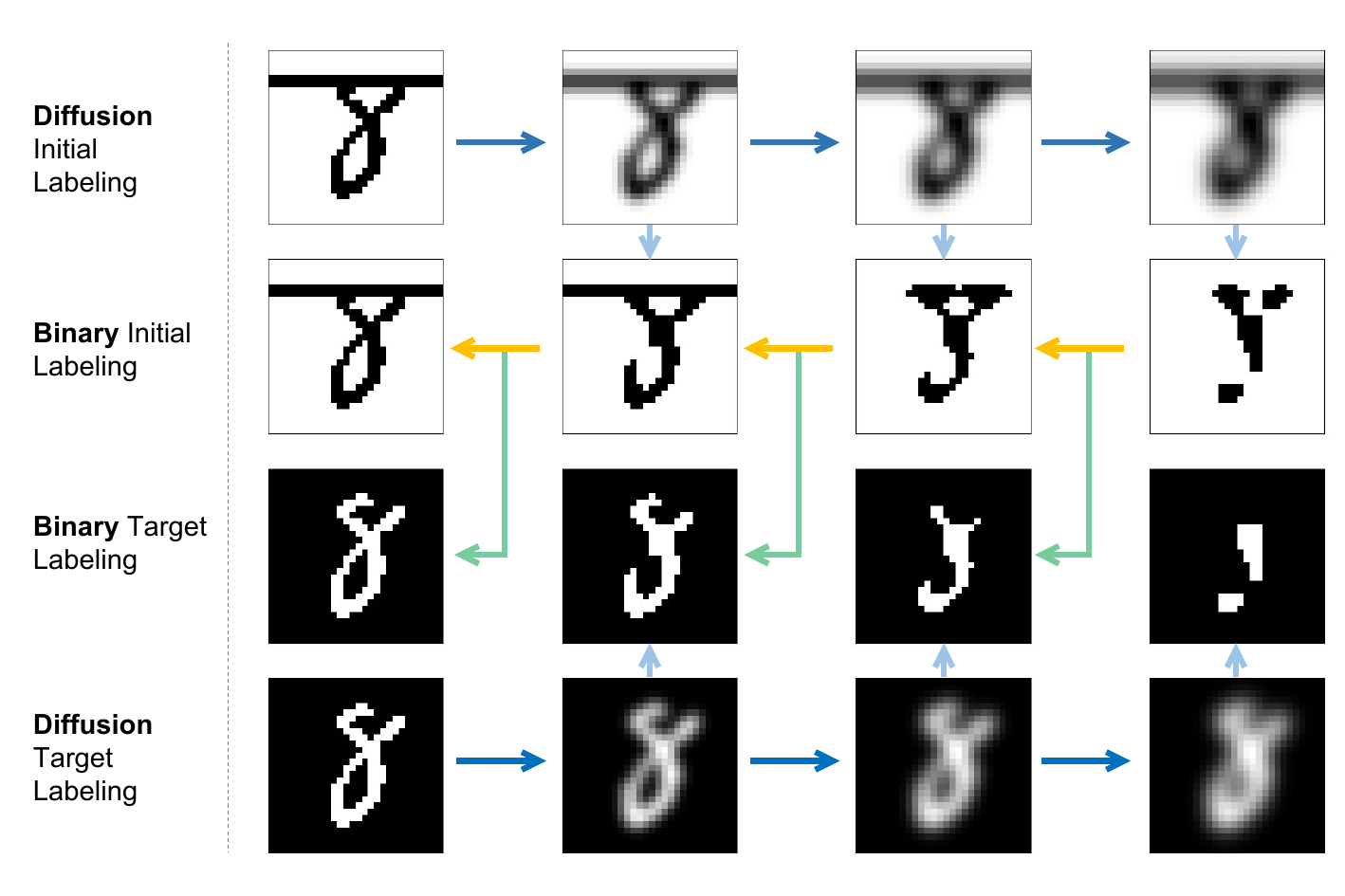}
        \caption{
        Image labeling via a linear scale space, exemplified for a corrupted handwritten digit ``8''. \textbf{Top row:} Scale-space representations $v_{1}\mapsto G(\cdot,\tau_{d})\ast v_{1},\; d\in[D]$, of the input data. \textbf{Bottom row:} In the training phase, analogous representations of the \textit{target} labeling $v^{\ast}_{T+1}\mapsto G(\cdot,\tau_{d})\ast v^{\ast}_{T+1}$ are provided as well. \textbf{Center rows, inference:} The \textit{learned} sequence of parameter matrices $\Omega_{t},\,t\in[T]$ enables the \textit{binary assignment flow network} \eqref{eq:def-binary-AF-network} to infer the uncorrupted labelings (3rd row; green arrows) from the corrupted labelings (2nd row; yellow arrows) across the scale space.
        }
     \label{fig:diffusion_schema}
\end{figure}

We supply the triangular subgraphs with a \textit{sequence} of initial data corresponding to convolutions with the Gaussian kernel
\begin{equation}
G(x,\tau) = \frac{\tau}{4\pi} e^{-\frac{\tau}{4}\|x\|^{2}},\qquad
\tau_{1} <\dotsb < \tau_{D}
\end{equation}
and an increasing sequence of parameter values corresponding to a \textit{coarse-to-fine} transition of spatial scale and structure; see Figure \ref{fig:diffusion_schema} for an illustration. The approach from Section \ref{sec:Approach-3x3} is applied to \textit{each} transition $\tau_{d}\to\tau_{d+1}$. The changing target labels induced by the linear scale space provide information about the \textit{spatial context at varying scales}, which facilitates both learning and inference.

Thus, this process is structured by two nested parameter sequences: The assignment flow time (network layer index) $t \in \{1,...,T+1\}$ for each transition $\tau_{d}\to\tau_{d+1}$ and the linear scale space parameters $d \in \{1,..,D\}$.
Accordingly, the loss \eqref{eq:loss-function} is extended to
$
    \mc{L} =  \sum_{d=1}^D \mc{L}(\xi_d).
$

\label{sec:Approach-ScaleSpace}

\section{Experiments and Discussion}\label{sec:Experiments}

\subsection{Data Sets}\label{sec:Data-Sets}

\begin{figure}
    \centering
    \begin{minipage}[v]{0.325\textwidth}
        \centering
        \includegraphics[width=\linewidth]{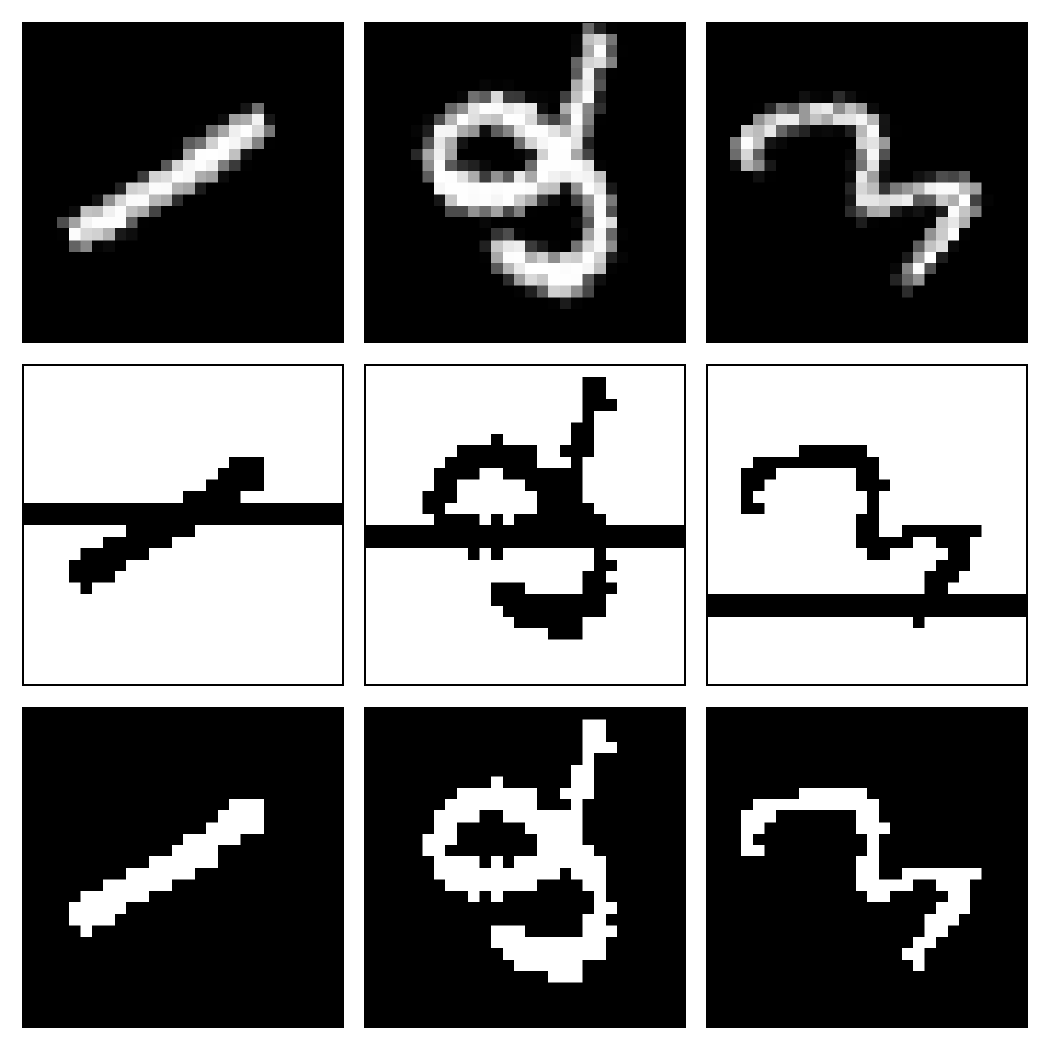}
    \end{minipage}
    \begin{minipage}[c]{0.433\textwidth}
        \centering
        \includegraphics[width=\linewidth]{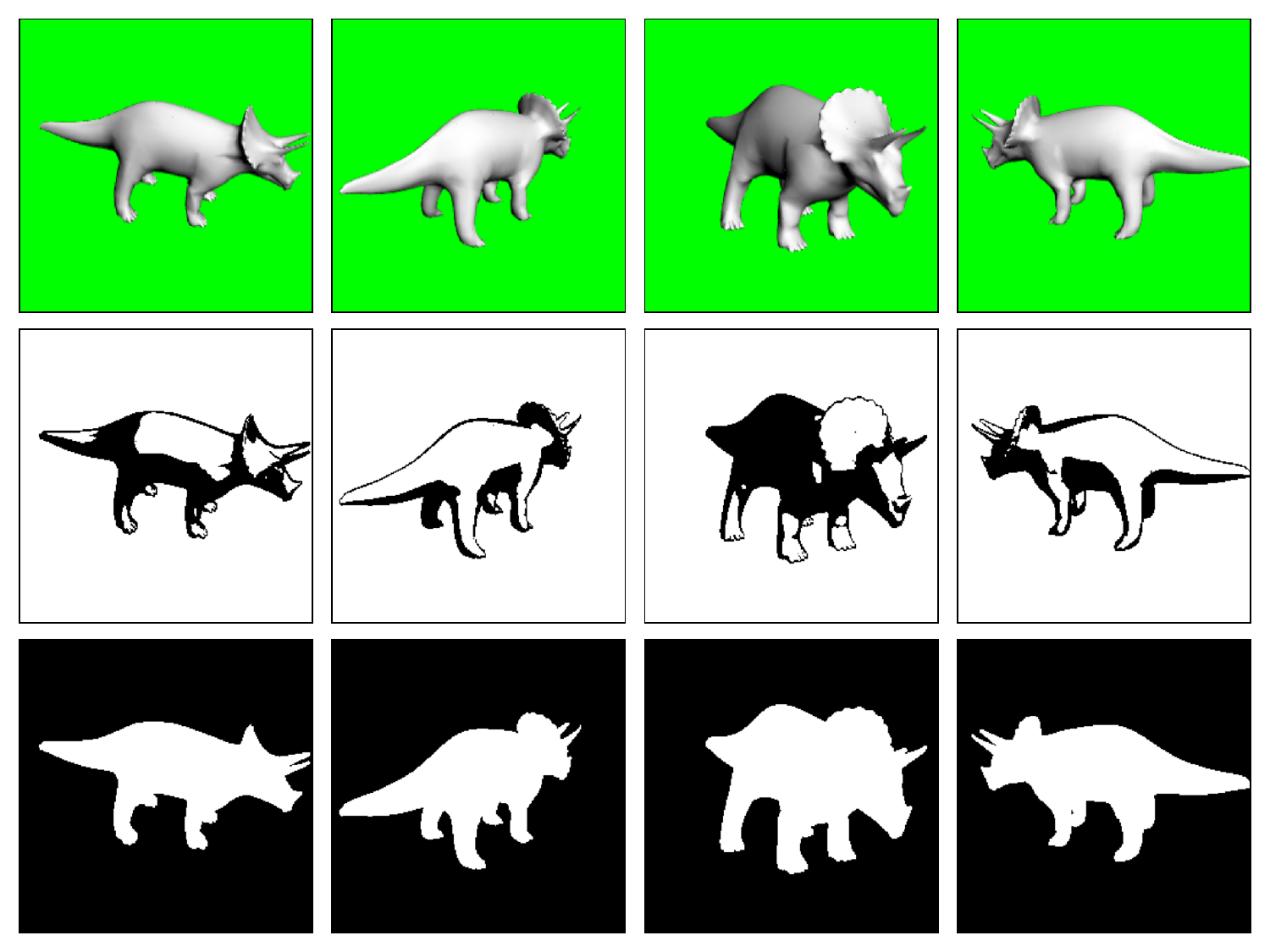}
    \end{minipage}
    \caption{\textbf{Left}: Examples of input images (top) corrupted initial labelings (mid) and target labelings (bottom) from the \textit{Rotated MNIST} data base. \textbf{Right}: Examples of generating images, initial labelings and target labelings for the Dino scene.}
    \label{fig:example_paper}
\end{figure}

We used the following data sets illustrated by Figure \ref{fig:example_paper}.
\begin{description}
\item[Rotated MNIST:] a publicly available well-known data set \cite{Larochelle:2007aa} which generalizes the original MNIST data set. Data points were \textit{additionally corrupted} (i) by selecting randomly two rows containing foreground pixels and changing all pixels in these rows to foreground; (ii) by interchanging fore- and background pixels afterwards. Thus the task is to \textit{simultaneously} invert fore- and background and to correct false labelings.
\item[Dino scene:] Images of computer-generated 3D scenes, turned into binary input labelings by thresholding. The task is to infer the foreground regions just from the irregular black-/white regions and the spatial context provided by the entire graph and local-to-global information aggregation, \textit{without} using explicit shape information as provided e.g.~by the silhouettes. This experiment mimics the famous psychophysical Dalmation dog experiment \cite[pp.20]{Frisby:1980aa} which demonstrates the ability of the human visual system to infer object shapes from seemingly unorganized binary spatial patterns. 
\end{description}

\vspace*{-0.75cm}

\vspace*{-1.0cm}
\subsection{Implementation Details}\label{sec:Implementation-Details}

\textbf{Network architecture.} The network architecture comprises 186.995 trainable parameters.
The temporal branch processes the time-dependent features via an LSTM cell with 56 hidden units, followed by a linear projection to 56 dimensions.
The static branch computes a 56-dimensional embedding of the input through a multilayer perceptron with hidden dimensions 256 and 512 and SiLU activations.
The two embeddings are added elementwise and passed through a 56-dimensional fusion layer and a final linear layer predicting the three $SO(2)$ rotation angles in \eqref{eq:network-F}. Since inference is done across the linear scale space (Section \ref{sec:Approach-ScaleSpace}), the corresponding scale index $d\in[D]$ was used as additional input datum.

\begin{figure}
    \centering
     \begin{minipage}[v]{0.3\textwidth}
        \centering
        \includegraphics[width=\linewidth]{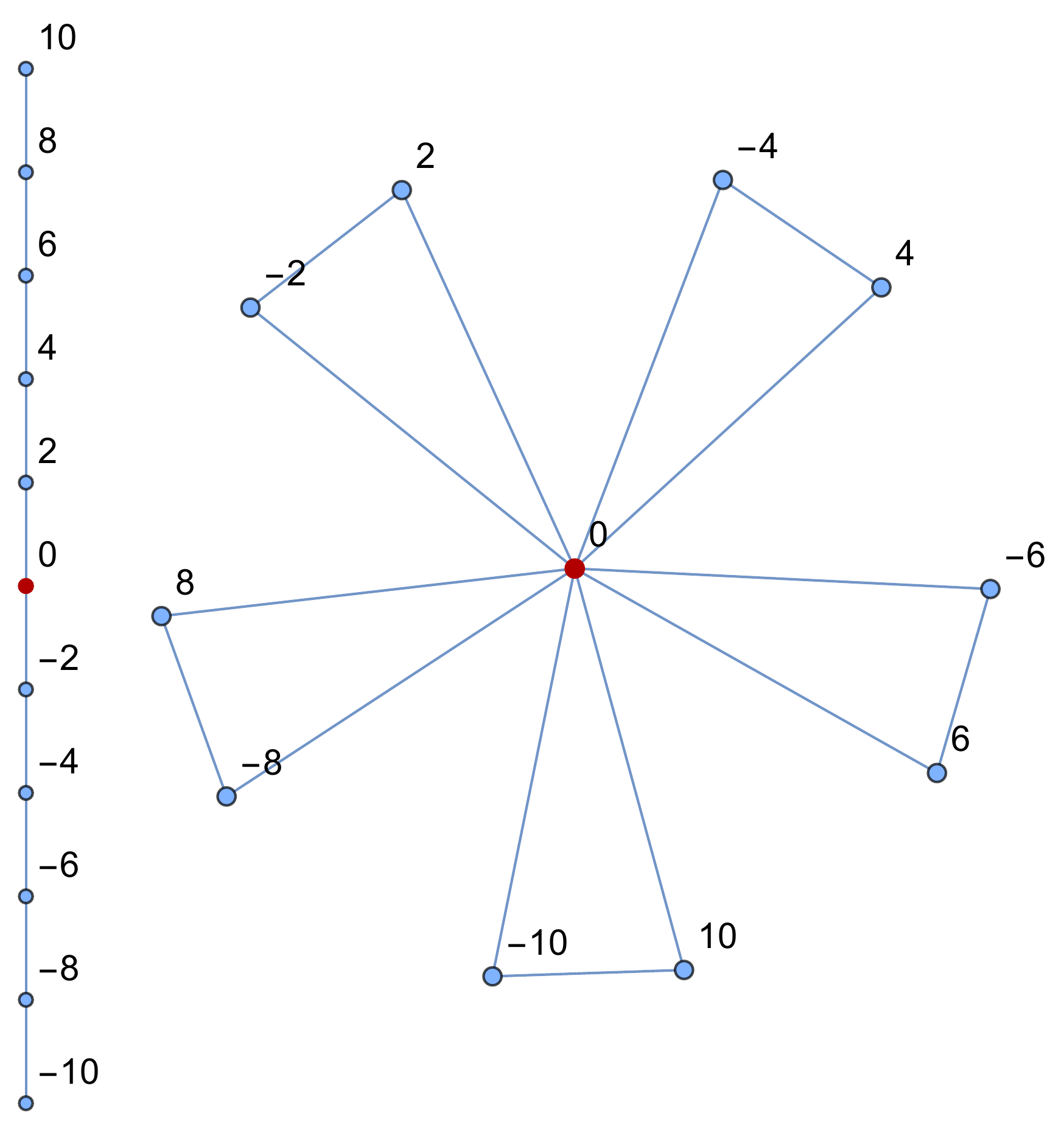}
    \end{minipage}
    \hspace{0.01\textwidth}%
\begin{tikzpicture}[baseline=(current bounding box.center)]
    \draw[dash pattern=on 2pt off 2pt, line width=0.5pt]
        (0,0) -- (0,3.5);
\end{tikzpicture}%
\hspace{0.01\textwidth}%
    \begin{minipage}[c]{0.6\textwidth}
        \centering
        \includegraphics[width=\linewidth]{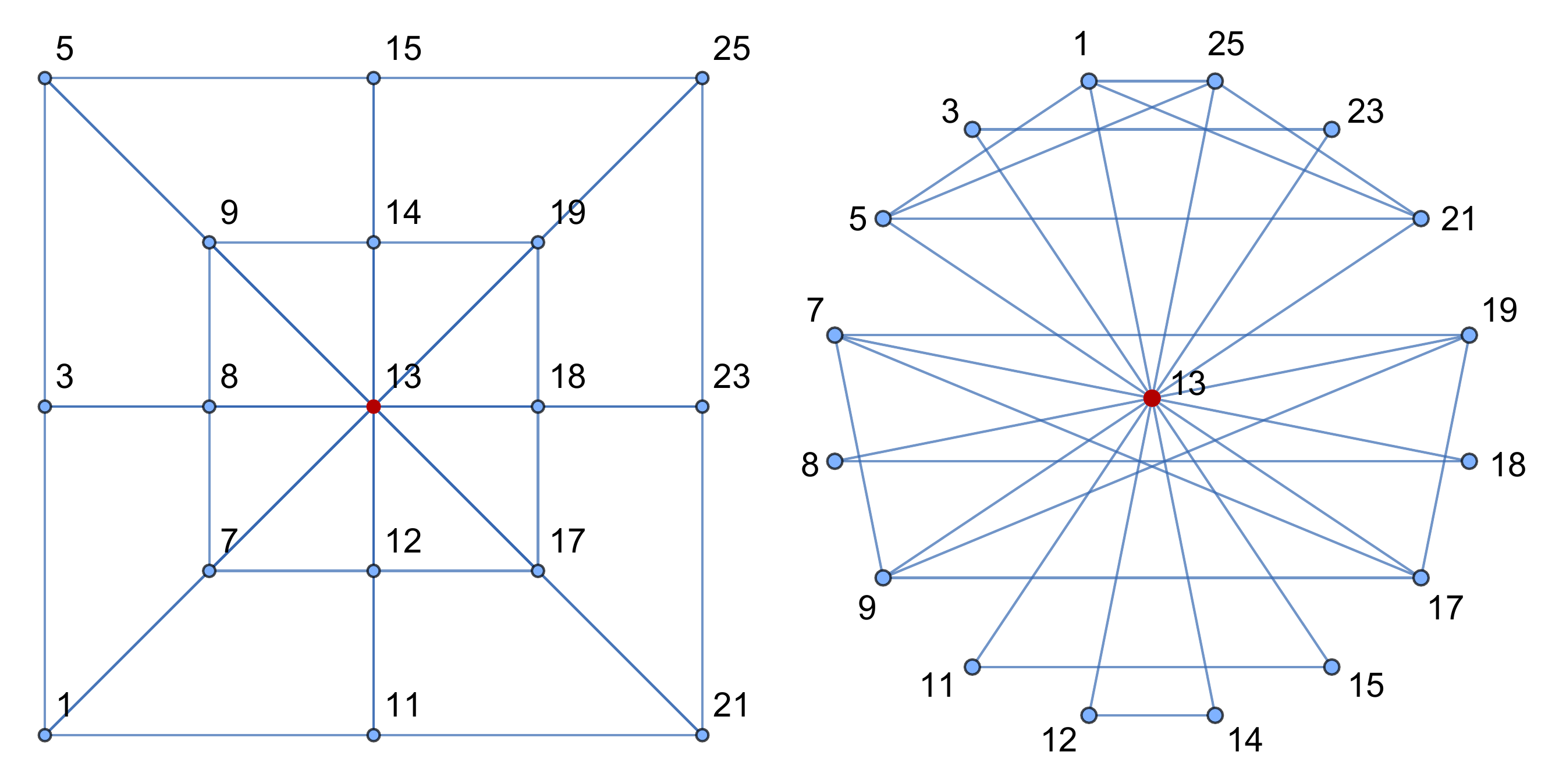}
    \end{minipage}
    \caption{
        \textbf{Triangular subgraphs.} \textbf{Rotated MNIST} (leftmost 2 panels): Vertices associated with a center vertex (marked red, left panel), used to define triangular subgraphs (right panel, different vertex embedding for visualization). \textbf{Dino scene} (rightmost 2 panels):  Vertices associated with a center vertex (marked red, left panel), used to define triangular subgraphs (right panel, different vertex embedding for visualization). For both scenes, every vertex becomes a center vertex (red) and thus defines a local collection of triangular subgraphs.
}
    \label{fig:triangle_configs}
\end{figure}
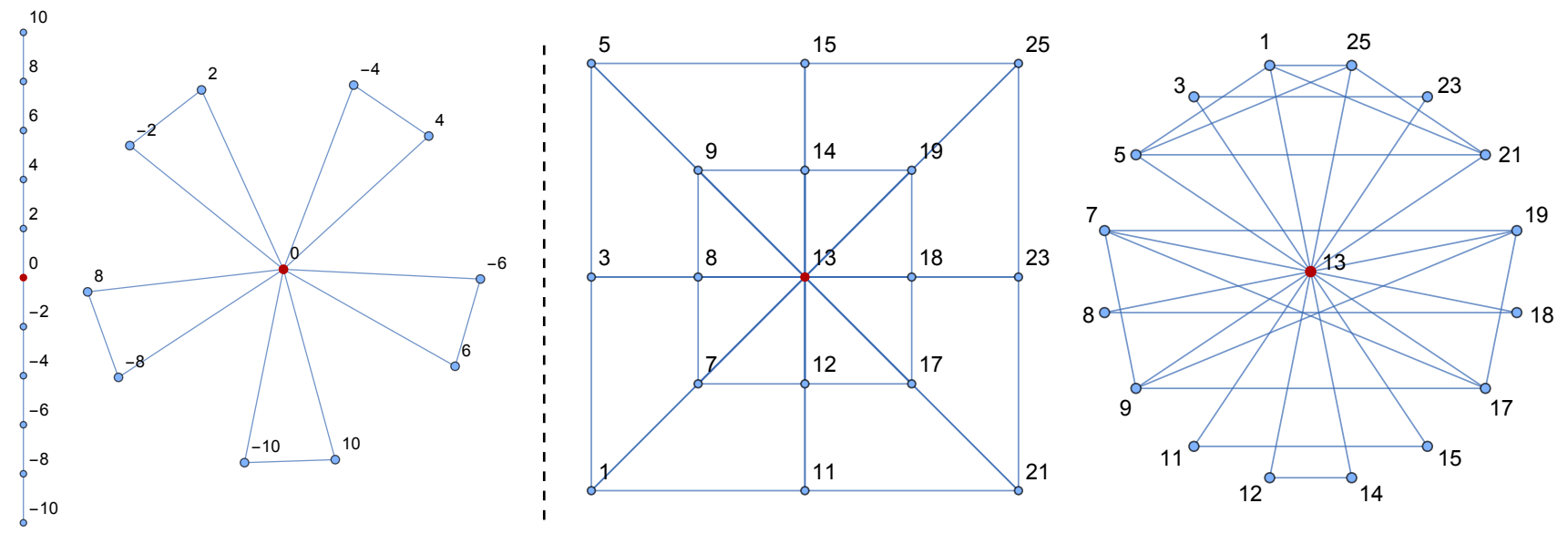

\textbf{Graph structure.} The set of edges added to the underlying grid graph in order to form \textit{triangular} subgraphs, was manually determined for each data set in an ad-hoc way in view of the inference task, as specified by Figure \ref{fig:triangle_configs}. Ideally, a suitable structure for the problem class at hand should be a \textit{variable} and adaptively inferred as well, which is beyond the scope of this paper and an attractive perspective for future work.

\newpage
\vspace*{-1.0cm}
\textbf{Training.} The model for Rotated MNIST was trained on 500 images (50 per digit class) of size $28 \times 28$, with 2.240 triangles complementing  the corresponding grid graph (cf.~Fig.~\ref{fig:triangle_configs}). We used three diffusion steps. 
As for the Dino scene, 140 images downsampled to $64 \times 64$ were used for training, with 61.752 triangles complementing the underlying grid graph.
We used four diffusion steps.
The computation was done on a desktop computer with a NVIDIA GeForce RTX 5070 Ti offering 16GB of VRAM.

The training time for the Rotated MNIST model was 8:58min for 45 epochs (all images are processed within one epoch).
Training the model on the Dino dataset took 29:30min for 70 epochs.  
Applying the Rotated MNIST model to a test image takes 0.02 sec, whereas the model trained for the Dino scene needs 0.1 sec for inference.

\vspace*{-0.5cm}
\begin{table}
\centering

\begin{minipage}[t]{0.49\textwidth}
\vspace{0pt}
\centering
\begin{tabular}{|r|r|r|}
\hline
\multicolumn{3}{|c|}{\textbf{Rotated MNIST}} \\
\hline
Diff.~Stage & Foreground Acc. & Background Acc. \\
\hline
0 & 97.26 \% & 98.83 \% \\
1 & 96.67 \% & 98.62 \% \\
2 & 92.83 \% & 98.34 \% \\
\hline
\end{tabular}
\end{minipage}
\hfill
\begin{minipage}[t]{0.49\textwidth}
\vspace{0pt}
\centering
\begin{tabular}{|r|r|r|}
\hline
\multicolumn{3}{|c|}{\textbf{Cow/Dino}} \\
\hline
Diff. Stage & Foreground Acc. & Background Acc. \\
\hline
0 & 97.35 \% & 99.90 \% \\
1 & 97.32 \% & 99.93 \%\\
2 & 94.02 \% & 99.80 \%\\
3 & 91.32 \% & 99.59 \%\\
\hline
\end{tabular}
\end{minipage}

\vspace{0.2cm}

\caption{\textbf{Prediction performance} for independent test data for Rotated MNIST (left) and the Dino scene (right). Recall from Fig.~\ref{fig:diffusion_schema} that during inference, the sequence of diffusion stages is $2\to 1\to 0$. The accuracy measurements have been averaged over the entire test data set. %
}
\label{tab:mnist_diffusion_results}
\end{table}

\vspace*{-1.0cm}
\textbf{Testing.} 
We computed the minimum Hamming distance between each predicted test labeling $X^{\text{test}}$ and \textit{both} the ground truth test labeling and the
closest labeling in the training set $\mathcal{D}_{\text{train}}$. %
For Rotated MNIST, the predicted test labelings differ from the closest training labelings on average at 10.25\% of the vertices, and by 2.74\% for the Dino data set.
This shows that the model generalizes to previously unseen data rather than merely reproducing labelings from the training set. Table \ref{tab:mnist_diffusion_results} reports the prediction performance for the independent test sets.

\begin{figure}[]
    \centering
    \begin{minipage}[v]{0.497\textwidth}
        \centering
        \includegraphics[width=\linewidth]{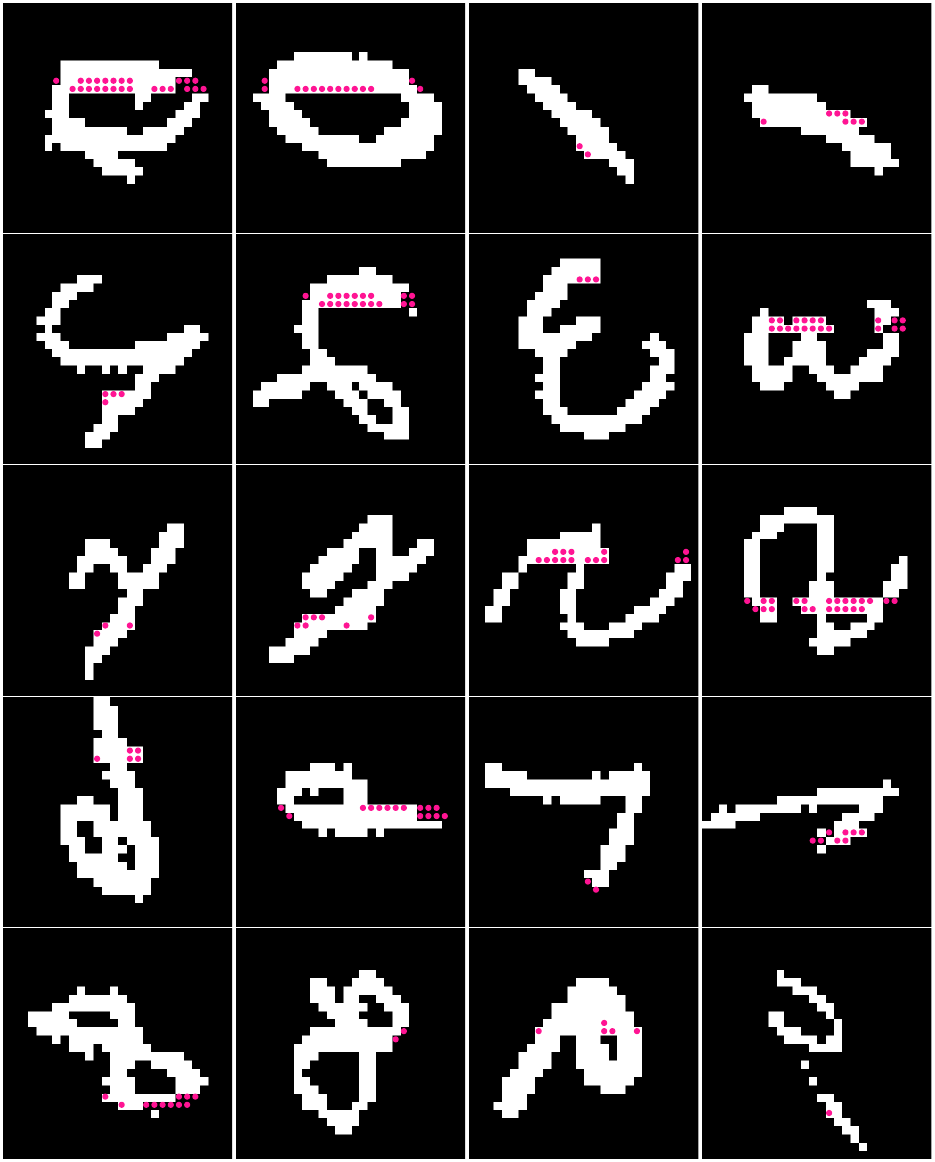}
    \end{minipage}
    \begin{minipage}[c]{0.497\textwidth}
        \centering
        \includegraphics[width=\linewidth]{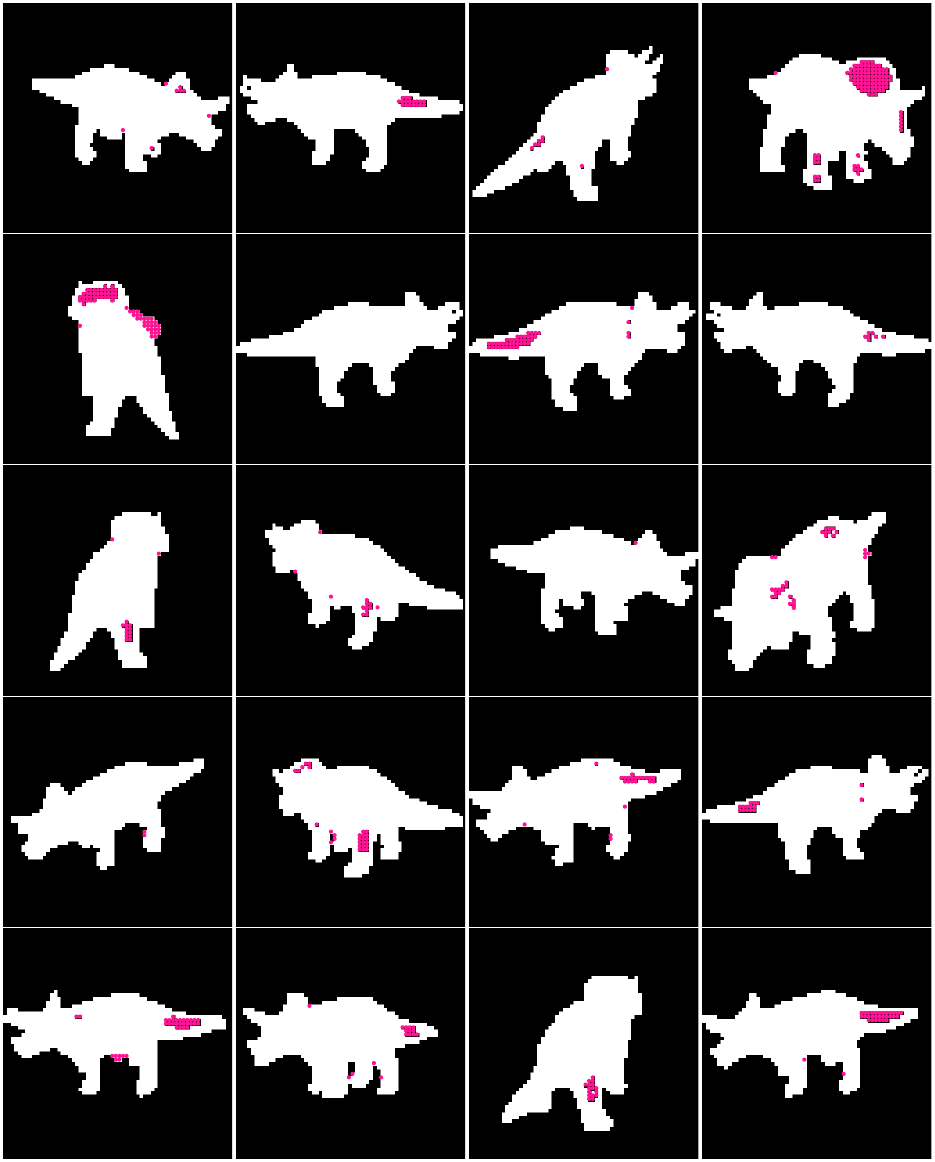}
    \end{minipage}
    \caption{Predicted labelings for independent test data for
        Rotated MNIST (left)
        and the Dino scene (right). Pink pixels indicate vertices with wrong label assignments.}
    \label{fig:Cow/Dino}
\end{figure}
\section{Conclusion and Outlook}\label{sec:Conclusion}
%

We introduced \textit{controlled} binary assignment flows based on geometrically fusing optimal parameter trajectories in \textit{low-dimensional} totally geodesic submanifolds. Our strategy of using \textit{cyclic} subgraphs for representing high-dimensional point in the symmetric space \eqref{eq:def-Om-general} complements established strategies using \textit{acyclic} subgraphs, or chordal subgraphs with low tree-width \cite{Vandenberghe:2014aa}, which may be of independent interest.

Regarding future work, the following topics are of particular interest.
\begin{itemize}
\item 
\textbf{Adaptive triangular subgraphs.} Ideally, the two subgraphical structures depicted by Figure \ref{fig:triangle_configs} should be learned from data, to enhance the adaptivity to different scenarios and to further improve prediction performance.
\item
\textbf{Adaptive scale space representation.} Providing non-local contextual information eases both learning and inference and enables to reduce the number of parameters. Using \textit{adaptive} scale selection and possibly \textit{non-linear} scale spaces might be beneficial in this respect.
\item
\textbf{Using symmetry.} The scenario `binary labeling' and our approach to using a symmetric space entails symmetries of the parameter representation (cf.~Figure \ref{fig:3x3-Traj-Averaged}), which have not yet been fully explored. Corresponding reductions might enable to consider large-scale scenarios efficiently regarding memory requirements and computation time.
\item
\textbf{Non-binary labelings.} The non-binary case requires various generalizations. Yet, a better understanding of the preceding points should pave the way for tackling arbitrary labeling scenarios in a similar way.
\end{itemize}

\begin{credits}
\textbf{Acknowledgements.} This work is funded by the Deutsche Forschungsgemeinschaft (DFG) under Germany's Excellence Strategy EXC-2181/1 - 390900948 (the Heidelberg STRUCTURES Excellence Cluster). This work is funded by the
Deutsche Forschungsgemeinschaft (DFG), grant SCHN 457/17-2, within the priority programme SPP 2298: Theoretical Foundations of Deep Learning.

\end{credits}

\vspace{-0.99em}
\bibliographystyle{splncs04}
\bibliography{TexInput/Controlled-AF.bib}

\end{document}